\documentclass[runningheads]{llncs}

\usepackage{graphicx}
\usepackage{comment}
\usepackage{amsmath,amssymb}
\usepackage{color}
\usepackage{url}
\usepackage{hyperref}
\usepackage{multirow}

\usepackage{subcaption}

\newif\ifreview
\reviewfalse

\ifreview
	\usepackage{lineno}

	\linenumbers
\fi

\begin{document}


\def\GCPRTrack{Special Track: Computer vision systems and applications}

\title{A Framework for Benchmarking Real-Time Embedded Object Detection}

\ifreview
\else

	
	\author{Michael Schlosser \and
	Daniel K{\"o}nig \and
	Michael Teutsch}
	
	\authorrunning{M. Schlosser et al.}
	
	\institute{HENSOLDT Optronics GmbH, Oberkochen, Germany\\
	\email{\{michael.schlosser, daniel.koenig, michael.teutsch\}@hensoldt.net}}
\fi

\maketitle              

\begin{abstract}
Object detection is one of the key tasks in many applications of computer vision. Deep Neural Networks (DNNs) are undoubtedly a well-suited approach for object detection. However, such DNNs need highly adapted hardware together with hardware-specific optimization to guarantee high efficiency during inference. This is especially the case when aiming for efficient object detection in video streaming applications on limited hardware such as edge devices. Comparing vendor-specific hardware and related optimization software pipelines in a fair experimental setup is a challenge. In this paper, we propose a framework that uses a host computer with a host software application together with a light-weight interface based on the Message Queuing Telemetry Transport (MQTT) protocol. Various different target devices with target apps can be connected via MQTT with this host computer. With well-defined and standardized MQTT messages, object detection results can be reported to the host computer, where the results are evaluated without harming or influencing the processing on the device. With this quite generic framework, we can measure the object detection performance, the runtime, and the energy efficiency at the same time. The effectiveness of this framework is demonstrated in multiple experiments that offer deep insights into the optimization of DNNs.

\end{abstract}

\section{Introduction}
Algorithm benchmarking is a crucial step during the development of computer vision systems and applications. In applications that have to meet requirements for minimum latency (i.e. real-time requirements) on limited hardware (i.e. edge devices), not only the algorithm effectiveness is important but also the efficiency. Facilitating recent state-of-the-art techniques such as deep learning and Deep Neural Networks (DNNs) as a very powerful tool in computer vision nowadays, authors try to find the sweet spot between processing speed and algorithm accuracy~\cite{Huang2017,Liu2022}. Relevant fields of application for finding such a trade-off range from robotics~\cite{Jung2018} and autonomous driving~\cite{Gog2021} to smartphones~\cite{Xiong2021}. Accompanying this progress, several vendors of specific inference hardware provide specialized software based optimization pipelines that enable scientists and developers to strongly reduce the latency of certain algorithms, while preserving the effective performance at the same time~\cite{intel_repo,nvidia_repo,vai_repo}. Since these optimization software frameworks and pipelines are highly specific, however, benchmarking certain rather generic computer vision algorithms for certain applications with related hardware limitations such as miniaturized edge devices can be challenging.

In this paper, we propose a generic framework for benchmarking low-latency computer vision algorithms for their application on vendor-specific inference hardware. We use object detection as our considered computer vision task. Utilizing the popular You Only Look Once (YOLO) object detector~\cite{Redmon2016} in its version YOLOv4~\cite{Bochkovskiy2020} as reference detection algorithm and MS Common Objects in Context (COCO)~\cite{Lin2014} as reference dataset, we aim to measure detection performance and algorithm runtime simultaneously. This is achieved by separating data distribution and evaluation from data processing. A host app running on a desktop computer distributes the video data via the light-weight Message Queuing Telemetry Transport (MQTT) protocol. This data is processed by a target app on the target board. The results are then sent back via MQTT to the host app for evaluation and benchmarking. In this way, we can efficiently compare different vendor-specific hardware and optimization software pipelines, which usually contain software tools for quantization and pruning of DNNs~\cite{Cai2017}. Furthermore, we can integrate new hardware and/or software updates easily into our framework to measure their performance and runtime gains, respectively.

Our proposed framework is not the first of its kind, of course. Several approaches exist for evaluating the performance of object detection, but most related literature either does not measure all important metrics relevant for the use on embedded systems~\cite{staecker2021,Wang2021}, which are primarily the accuracy, the runtime and the power consumption, or the proposed frameworks are less generic compared to ours as they only refer to a single hardware~\cite{staecker2021,Wang2021} or vendor~\cite{Rungsuptaweekoon2017}. Other authors compare embedded hardware for certain computer vision tasks but they do not use or mention a unified evaluation framework at all~\cite{Lin2019,verucchi2020,Yu2018}. Another relevant aspect is that our MQTT based publish-subscribe approach is more flexible than server-client based architectures~\cite{Rungsuptaweekoon2017}: in this way, we can evaluate a computer vision algorithm on multiple target platforms simultaneously. Server-client architectures instead usually have higher communication overhead since requests have to be sent to the host from each individual target device. Furthermore, MQTT is very light-weight, simpler, and with higher throughput compared to other protocols or interfaces such as Hypertext Transfer Protocol (HTTP) or Advanced Message Queuing Protocol (AMQP)~\cite{Gemirter2021,Guendogan2018,Mishra2021,Yokotani2016}. The application behind the framework is to provide a generic test bed for benchmarking optimized computer vision algorithms on highly efficient edge devices. In this way, we can effectively find the trade-off between processing speed and algorithm accuracy.

Our contributions are:
\begin{enumerate} 
\item We propose a generic evaluation framework for different embedded devices using a lightweight concept that can be deployed with little effort and that provides remote access from the host app without the need to send requests from the target side.
\item To the best of our knowledge this is the first framework of its kind that uses the highly efficient publish-subscribe protocol MQTT instead of less flexible or efficient communication protocols and/or interfaces such as HTTP or Representational State Transfer (REST). In this way, we can evaluate an algorithm on multiple target devices simultaneously and efficiently. 
\item We demonstrate the usefulness of our framework by evaluating two different embedded devices from two vendors utilizing their specific optimization pipelines for the task of generic object detection.
\end{enumerate}

The remainder of this paper is organized as follows: related work is presented in Section~\ref{sec:relwork}. Our proposed framework together with the related methodology is presented in Section~\ref{sec:method}. Experimental results are described in Section~\ref{sec:experiments} followed by a discussion in Section~\ref{sec:discussion}. We conclude in Section~\ref{sec:conclusion}.

\section{Related Work}
\label{sec:relwork}
Several approaches exist for evaluating the performance of object detection, but most proposed works either do not measure all important metrics for use on embedded systems, which are primarily the accuracy, the runtime and the power consumption, or are not generic enough, because they only refer to a single hardware or manufacturer.

Stäcker et al.~\cite{staecker2021} evaluate object detection on embedded systems using RetinaNet~\cite{Lin2017}. The detection model is optimized using Nvidia TensorRT and deployed on an Nvidia Jetson AGX Xavier. Accuracy and runtime, but not power consumption, are measured during the experimental evaluation. The Robot Operating System (ROS) is used to communicate with the target and to receive the detection results. Only an Nvidia Jetson AGX Xavier is used as hardware platform and the comparison to other platforms is not considered here. Rungsuptaweekoon et al.~\cite{Rungsuptaweekoon2017} utilize the object detector YOLOv2 on different embedded devices such as the Jetson TX1 and TX2. In addition to Frames Per Second (FPS) and mean Average Precision (mAP), the power consumption is also measured. A benchmark environment based on a server-client model is provided, where the target board acts as a client and requests the image data from a host PC. A system based on a client-server architecture is proposed, which requires the direct access of the target board to communicate with the server and the authors only consider Nvidia hardware.

Besides the evaluation of GPU-based hardware accelerators, there are also works dealing with the evaluation of object detection on Field Programmable Gate Arrays (FPGAs). Wang et al.~\cite{Wang2021} evaluate a YOLOv3 object detector, which is optimized using the recommended optimization framework Vitis AI on a Xilinx Zynq UltraScale+ MPSoC ZCU104 evaluation board. The performance results of the Xilinx MPSoC are compared to the performance of a Nvidia GeForce GTX 1080 GPU on a desktop computer. The power consumption and the FPS are measured, but no accuracy, which means that it is not verified whether the model still has an acceptable accuracy after optimization. Furthermore, there is no uniform evaluation framework for the different system environments that are compared, so implementation mismatches between the platforms are possible. In \cite{Yu2018}, Yu et al. propose a comparison of several object detectors such as YOLO, Faster RCNN and SSD, on multiple platforms, including Nvidia TK1, Xilinx Zynq 7045 and Xilinx KU115. They measure power consumption, throughput, and accuracy. However, they provide an incomplete comparison as not all models are evaluated on all boards. Furthermore, they also do not use an uniform evaluation framework for different implementations. Blott et al.~\cite{blott2021} propose a theoretical and experimental evaluation of different DNNs for several computer vision tasks on a variety of different acceleration platforms, such as FPGAs, GPUs and TPUs. The authors point out that measurement methods are often unclear and complicated by the large variety of deployment parameters. For this purpose, a large experimental study is conducted on how different DNN topologies behave with different deployment settings (e.g., batch size or power modes) and optimization methods (e.g., pruning and quantization), in terms of throughput, inference time, hardware utilization, power consumption, and accuracy.  Accordingly, a detailed study on the comparison of DNNs, especially in the use case of image classification, is presented here. Lin et al.~\cite{Lin2019} provide a benchmark for different SSD models deployed on an Intel Arria 10 FPGA for traffic sign detection. In addition to an analysis of the training framework, they evaluate the models deployed on the FPGA in terms of inference time, accuracy, and power efficiency, varying critical parameters such as floating point precision and batch size. They found that the inference time on the GPU is faster in most cases and that the FPGA is better in terms of power efficiency. As with most of the publications just presented, the focus here is an experimental evaluation of the results for the given models, optimization methods and hardware platforms rather than the methodology of a generic evaluation framework. Verucchi et al.~\cite{verucchi2020} propose a detailed comparison between object detectors, such as YOLOv3, CenterNet, and SSD deployed on an Nvidia Jetson AGX Xavier, a Xilinx Zynq UltraScale+ MPSoC ZCU102, and an industrial PC. High attention is paid to the fairness of the comparison between the detectors deployed on different hardware platforms. All important metrics  mentioned before are considered, which include accuracy, runtime, and power consumption. However, no uniform evaluation framework is implemented and the accuracy is directly measured on the target device.

\section{Proposed Evaluation Framework}
\label{sec:method}
Our proposed framework enables the evaluation of object detection algorithms for such a stream on arbitrary target hardware and shifts the responsibility of providing input data and calculating evaluation measures to a separate host PC. The framework itself, however, is generic and thus not limited to the task of object detection.
We choose an MQTT-based approach, because it is a light-weight protocol and often used in embedded devices~\cite{Mishra2021}.
Since MQTT uses a publish-subscribe mechanism, it is even possible to evaluate object detection on multiple target devices simultaneously. The MQTT broker manages message distribution and decouples the communication of host and target. Hence, requests do not have to be sent directly to the host from each individual target device as is the case with a commonly used server-client architecture.
\begin{figure}[htp]
\includegraphics[width=\textwidth]{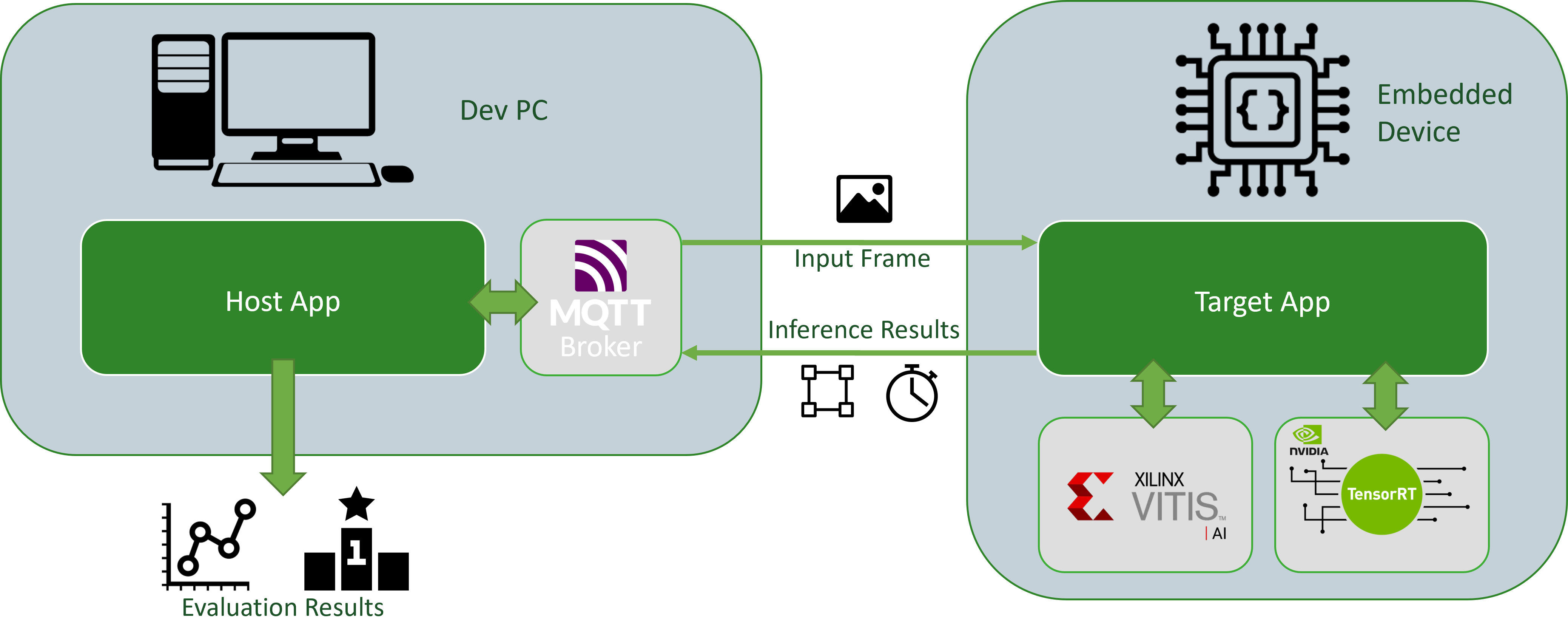}
\caption{Architecture of the proposed evaluation framework consisting of the development PC with the host application and the embedded device with the target application. MQTT serves as communication protocol between the devices.} \label{fig:eval-framework}
\end{figure}
As Fig.~\ref{fig:eval-framework} depicts, with the proposed evaluation framework image data is sent from a host computer to the target via MQTT. The target is the platform that performs the inference, the associated pre-processing, which consists of resizing and normalizing the input image data, and post-processing, which consists of the conversion of the model output to bounding box coordinates and confidence values, and Non-Maximum Suppression (NMS). The results are sent back to the host PC for evaluation.
The use of such a framework allows to evaluate a DNN directly on a platform that is used for deployment in operational systems. Thus, we are able to verify the conversion from common deep learning frameworks such as PyTorch or TensorFlow into the proprietary optimization frameworks of the hardware vendors such as Xilinx or Nvidia. In this work, without limitation of generality, we focus on Xilinx Vitis AI and Nvidia TensorRT. Furthermore, we can provide accuracy measures of a final deep learning application deployed on the embedded device. Various proprietary profiler tools are provided by Xilinx and Nvidia that support measuring of runtime and throughput. However, to evaluate complete accuracy measurements locally, the entire test dataset would have to be installed on the target. This can be impractical especially for large datasets such as the COCO test dataset that is used in this work and consists of over \mbox{20,000} images. We avoid this with the proposed evaluation framework. The responsibility of storing and providing the input data can thus be transferred to an external device (i.e. the host PC) together with the calculation of evaluation measures.

\begin{figure}[htp]
\centering
\includegraphics[width=0.96\textwidth]{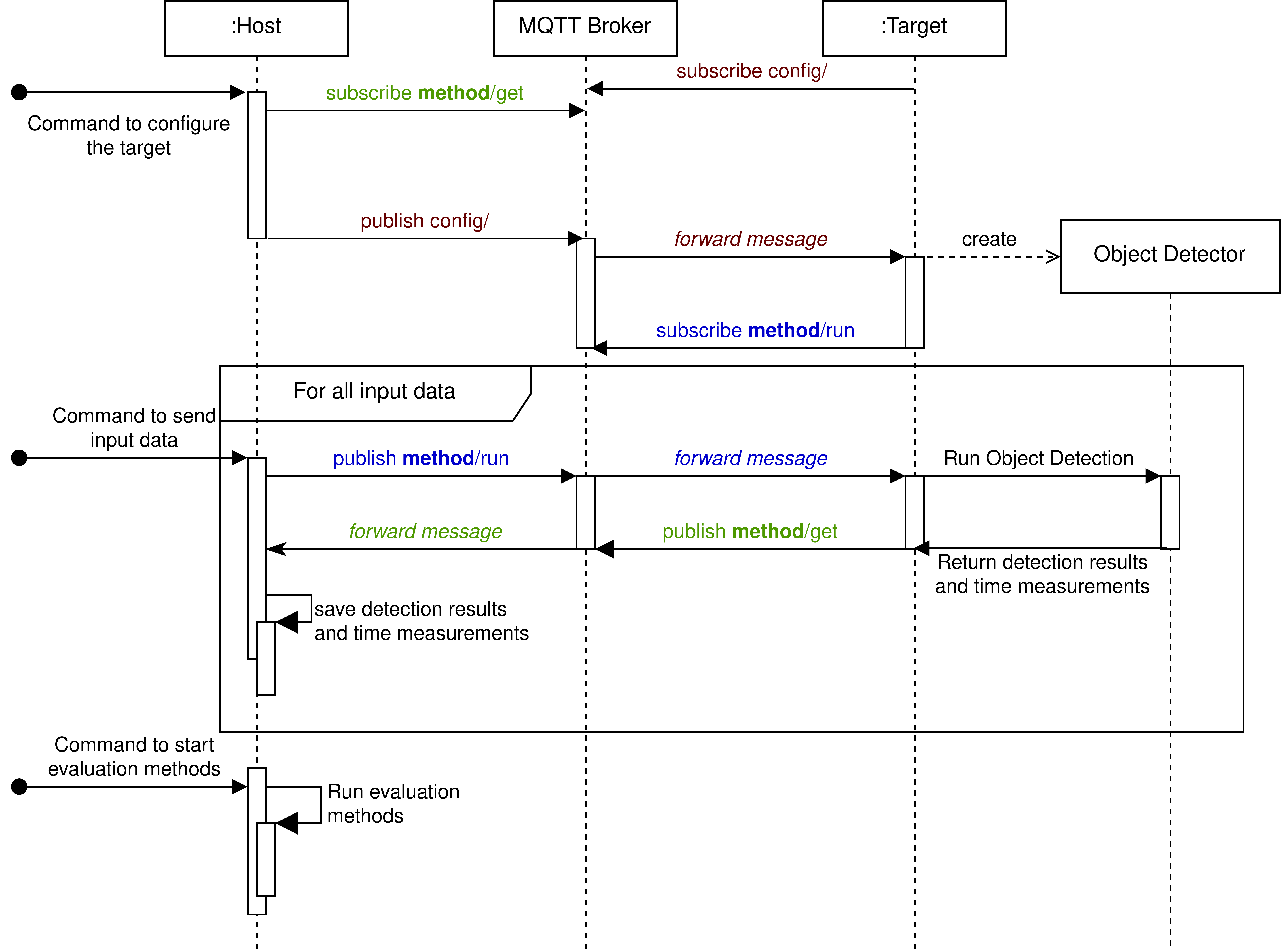}
\caption{UML sequence diagram for an evaluation session.} \label{fig:uml:sequence-diagram}
\end{figure}
Figure~\ref{fig:uml:sequence-diagram} depicts the flow of an evaluation session as a Unified Modeling Language (UML) sequence diagram. In the first preparatory step, the host sends the desired configuration of the experiment setup to the MQTT broker. The target subscribes to this configuration topic, whereupon messages are distributed that contain, for example, the algorithm to be used and the topics to be subscribed to, on which the input data is sent. This procedure simplifies the setup of the target, because the target component only has to be started, which can also be automated in the boot process, and all further steps can be done remotely from the desktop PC. The host app sends the input data to the MQTT broker in a next step. For this purpose, the input image is read in as an OpenCV~\cite{opencv_library} image matrix and serialized for transfer to the target device using Msgpack~\cite{Hamerski2018}. These input messages are then successively forwarded to the target component, processed there and the results are sent back. These results include the measured times of pre- and post-processing and inference, as well as the determined bounding boxes. For demonstration purposes, the processed image with the bounding boxes plotted can be sent back to the host via MQTT. The host can then display the processed images at runtime if required. After all desired input data has been processed, the results can be further processed using the selected evaluation modules. For example, the received detection results can now be compared with the annotations of the test dataset that was just processed in order to measure the accuracy.

\begin{figure}
\includegraphics[width=\textwidth]{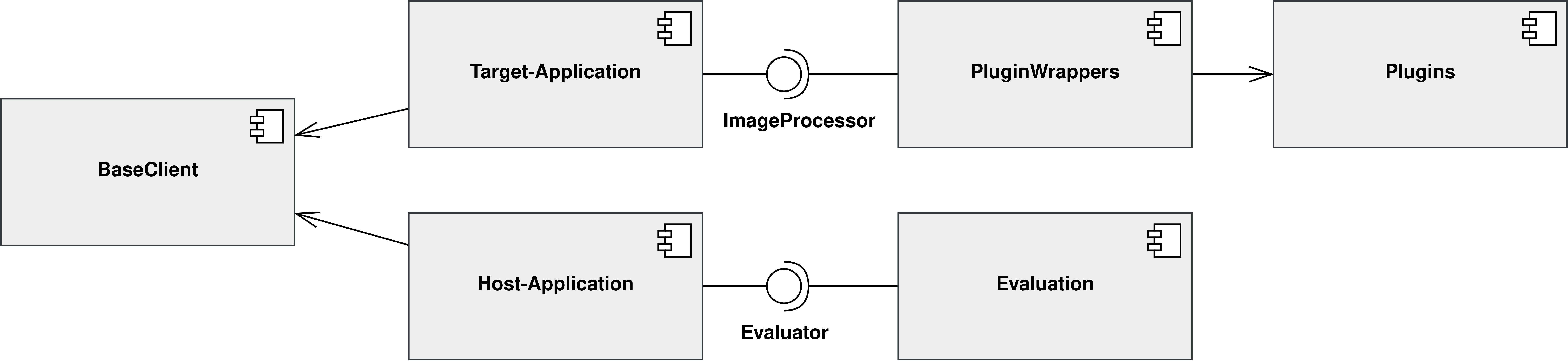}
\caption{UML component diagram for the evaluation framework.} \label{fig:uml:component-diagram}
\end{figure}
Due to the decoupling and the modularized structure of the framework, as Fig.~\ref{fig:uml:component-diagram} depicts, it is possible to integrate further object detection algorithms or evaluators without much effort. Hence, no additional dependencies arise for the communication of the two components via MQTT and the actual object detection application. Functionality relevant to both target and host such as data transfer in particular, is implemented in the BaseClient component. By implementing the evaluator interface, a plugin can be developed for the host component that has access to the result message. This allows further processing of all existing measurements. Similarly, the object detector interface can be implemented on the target board, which receives and processes an OpenCV image matrix and provides access to various information such as the predicted bounding box coordinates and time measurements.
The target part of the evaluation framework only depends on an MQTT library, such as Eclipse Paho MQTT~\cite{pahomqtt_repo} and Msgpack, which is used for the standardization and de/serialization of the message. All other dependencies are dependent on the respective plugin. For example, the Vitis AI Runtime must be installed as a dependency for controlling an application for the Xilinx Deep Learning Processor Unit (DPU), which is a programmable engine for accelerating deep learning tasks on an FPGA~\cite{dpu-guide}. Since host and target are completely decoupled from each other, modifications to the host component are not necessary when including a new target device.

\section{Experiments}
\label{sec:experiments}
In this work, we choose a YOLOv4 object detector~\cite{Bochkovskiy2020} as baseline model. For the Xilinx Zynq UltraScale+ MPSoC ZCU104~\cite{zcu104-guide} it is optimized using Xilinx Vitis AI~\cite{vai_repo} and for the Nvidia Jetson AGX Xavier~\cite{jetson-guide} we utilize Nvidia TensorRT~\cite{nvidia_repo}. By varying different parameters within the vendor-specific optimization frameworks, we can analyze the effects on the then deployed DNN. This for example includes steps such as quantization, which is the approximation of a neural network in floating point precision by means of a new neural network with a smaller bit width~\cite{Gholami2021}, and pruning, in which the size of the neural network is reduced by the systematical removal of elements~\cite{Blalock2020}. In addition, a cross-platform comparison between the models on different target devices is performed.

\subsection{Experimental Setup}
\begin{figure}
\includegraphics[width=\textwidth]{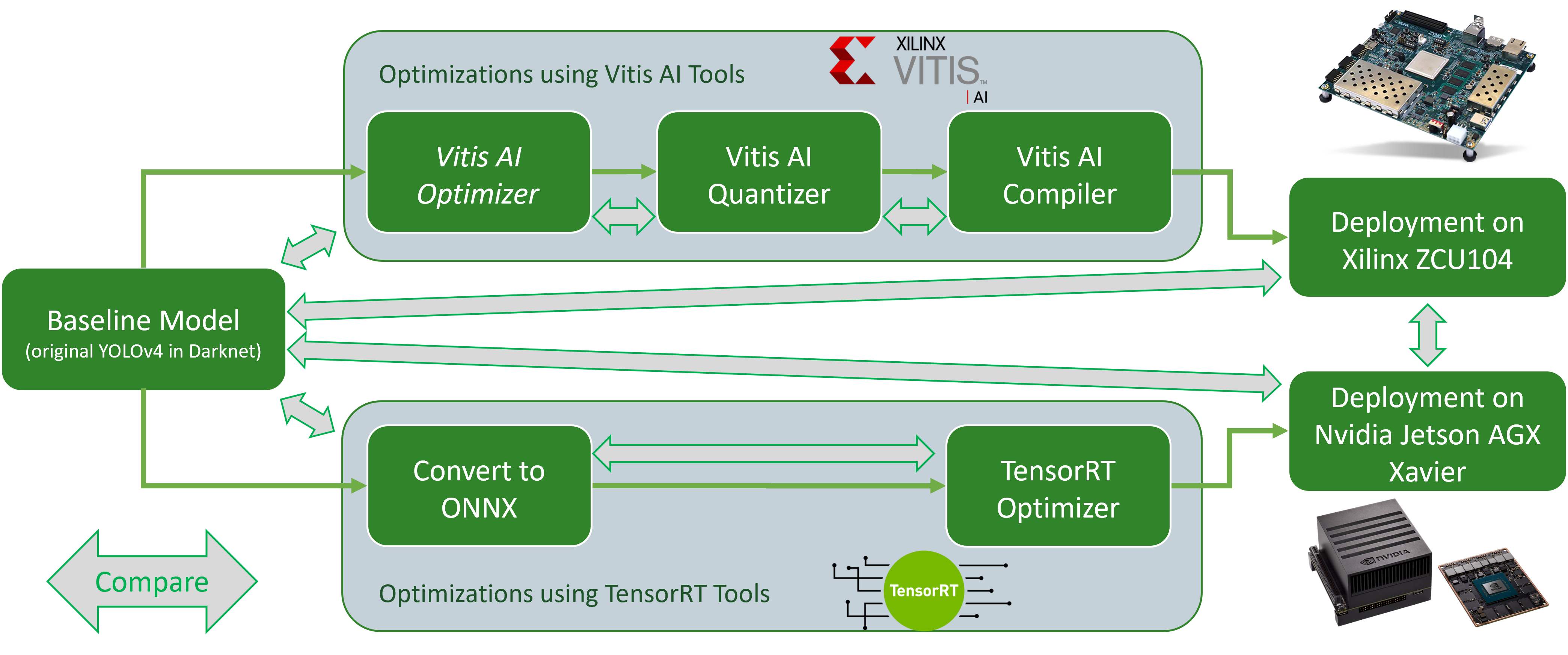}
\caption{Optimization pipelines for Xilinx Vitis AI and Nvidia TensorRT.}
\label{fig:opt-pipeline}
\end{figure}

In order to achieve comparability between the optimization frameworks, we set up the conversion and optimization process as visualized in Fig.~\ref{fig:opt-pipeline}.
We select a model trained with a deep learning framework compatible with both TensorRT and Vitis AI: TensorFlow. Accordingly, the first step is to convert a YOLOv4 model to TensorFlow, which was originally trained in Darknet~\cite{darknet13}, to create a baseline for the optimization pipeline.

Vitis AI provides three separate optimization tools. The Vitis AI Optimizer, which applies the pruning, the Vitis AI Quantizer, which quantizes the model into INT8-precision and the Vitis AI Compiler, which compiles the model for the underlying DPU design.
Pruning with the Vitis AI Optimizer is not considered in the optimization pipeline directly, since the related software tool is not free of charge. Its effects are nevertheless analyzed in a separate experiment. If the validation results are satisfactory and only a slight or even no loss of accuracy can be noted, the quantized model must be compiled using the Vitis AI compiler.

For optimization with TensorRT, we convert the TensorFlow model into ONNX format as a first step. For this purpose, the command line tool \mbox{onnx-tf} \cite{onnx-tf_repo} is used.  Subsequently, the ONNX model is optimized for utilization on inference hardware using TensorRT. According to the official documentation of the tool, \mbox{onnx-tf} not only provides a portable intermediate format that increases the interoperability of the model, but it also performs optimization steps during the conversion that are intended to reduce the inference time. Pruning is already performed in this step: unused ONNX operators (e.g., neurons of the DNN) are removed and similar operators are merged.
The TensorRT Optimizer processes the converted ONNX model. Therefore, the optimizer must be executed directly on the target hardware, since various hardware-specific optimizations are performed.

As evaluation measures, we choose the inference time, FPS, the energy consumption in watts, the energy efficiency in FPS/watt, and the mAP 0.5:0.95 for the accuracy.
For the measurement of the inference time, the object detector plugins were provided with time measurements, which are sent to the host along with the result message.
The measurement of power consumption in watts is realized by reading a multimeter, which is connected in series with the hardware platform to be evaluated. To measure the power in watts, the constant set voltage is multiplied by the current. A distinction is made between absolute and relative power consumption during the execution of object detection. To determine the relative power, the idle power, which is the power consumption of the target device without running an application, is subtracted from the absolute power used to actually execute the application. This increases the comparability between the inference hardware since the absolute power of the two platforms differs greatly as shown in Table~\ref{tab:results}.
Energy efficiency indicates the FPS in relation to the power consumption. This enables us to analyze how efficiently the resources of the inference hardware are used in terms of power consumption. Due to the diversity of the compared hardware in terms of computing power and power consumption, measuring the energy efficiency allows for a cross-platform comparison, which puts the energy consumption of the platform in relation to the runtime performance.
Accuracy is measured by averaging the mAP for Intersection-over-Union (IoU) thresholds from .5 to .95 in steps of .05. mAP is a standard metric for evaluating object detectors and used for the COCO benchmark~\cite{Lin2014}.

\subsection{Implementation Details}
The baseline model is available in resolutions of $416 \times 416$, $512 \times 512$, and $608 \times 608$ pixels. It was found that the $512 \times 512$ resolution provides the best compromise between accuracy and runtime. Therefore, this model is used as the baseline model for the majority of the experiments. Nevertheless, in a separate experiment (see Fig.~\ref{fig:map-vs-infer},~\ref{fig:map-vs-eff} and \ref{fig:map-vs-watt}), the other two variants are also evaluated and included in the final comparison.
All executions of the object detection algorithm are performed with similar pre- and post-processing, which are based on reference implementations of the respective framework. For example, for Vitis AI, the pre- and post-processing is implemented following the example of the Vitis AI Library, which is a highly optimized variant. For TensorRT and ONNX, pre- and post-processing is implemented via Python numpy. For a fair comparison, we verify that all implementations are similar across all frameworks.
For each evaluation session, object detection with the COCO validation dataset consisting of 5,000 frames of image data is run six times in total and the evaluation results are averaged to ensure more stable results in terms of runtime. For the measurement of the accuracy, the confidence threshold is uniformly set to 0.25. For the execution of the NMS in post-processing, a threshold of 0.45 was set. The choice of threshold values was taken from existing implementations of a YOLOv4 object detection application~\cite{repo_yolo_tf_converter,vai_repo}.
Accuracy measurements on the test dataset, consisting of about 20,000 frames of image data, are performed only once.
The annotations of the test dataset are not freely available, since they are used for participation in the COCO challenge, and the related website allows to upload results to the COCO server just up to five times a day. In the context of this work, accuracy measurements have been performed with the test dataset for each optimization level in this way. For optimizations that have no influence on the accuracy, only the validation dataset, consisting of 5,000 frames of image data, is entered several times for the evaluation, since the annotations of the validation dataset are freely available and the accuracy can thus be determined using the proposed evaluation framework.
The confidence threshold was reduced to .05 for the measurements on the test dataset, since this setting is used to evaluate the original darknet YOLOv4. It is a common practice to set the confidence threshold for evaluating the accuracy that low to achieve a higher mAP~\cite{yolov3}.

\subsection{Experimental Results}
For the evaluation of the TensorRT optimizer, the model was quantized to all possible precisions such as Integer representations using 8 Bits (INT8) and floating point representations using 32 Bits (FP32) and 16 Bits (FP16), respectively.
\begin{figure}[htp]
    \centering
    \begin{subfigure}{0.49\textwidth}
        \includegraphics[width=\textwidth]{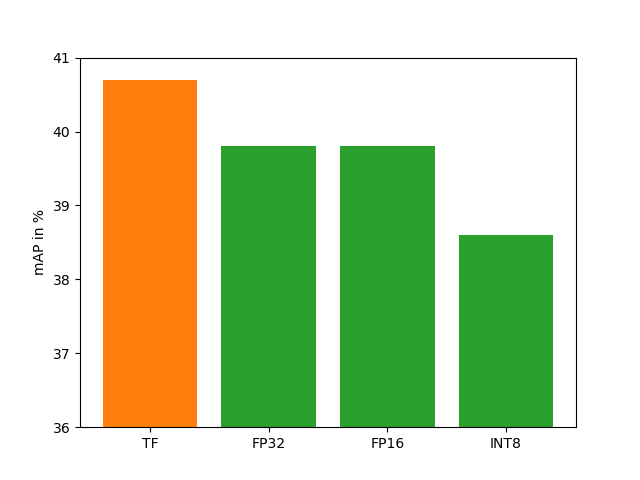}
        \caption{TensorRT}
        \label{fig:map-trt}
    \end{subfigure}
    \hfill
    \begin{subfigure}{0.49\textwidth}
        \includegraphics[width=\textwidth]{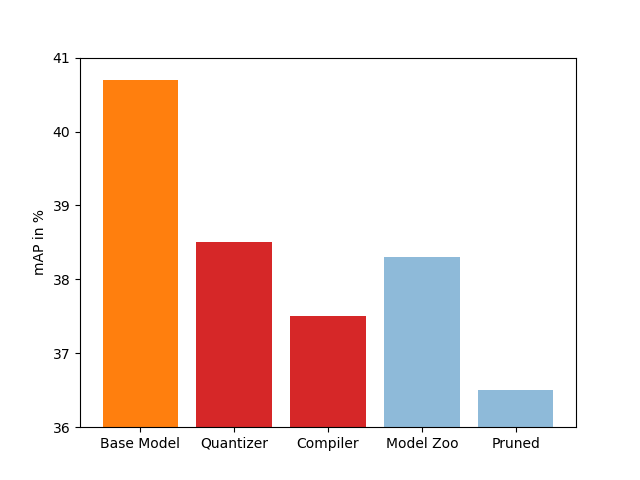}
        \caption{Vitis AI}
        \label{fig:map-vai}
    \end{subfigure}
    \caption{Accuracy comparison of optimized DNN models after all optimization steps. A higher bar indicates higher mAP and thus better DNN performance. TF is the reference DNN run in TensorFlow at FP32 precision.}
\end{figure}

Figure~\ref{fig:map-trt} shows the accuracy loss when quantizing to all available precisions. It could be observed that optimization with TensorRT in FP32 format results in accuracy losses. However, when quantizing the model converted to FP16 format, the same effects on the accuracy could be observed. But the inference time could be almost halved compared to the FP32 variant as we show in Table~\ref{tab:results}.
\begin{table}
    \centering
    \renewcommand{\arraystretch}{1.3}
    \begin{tabular}{c|c|c|c|c|c|c|c}
        \multirow{2}{*}{\textbf{Framework}} & \multirow{2}{*}{\textbf{Device}} & \multirow{2}{*}{\textbf{Precision}} & \textbf{Inference} & \textbf{Absolute} & \textbf{Relative} & \multirow{2}{*}{\textbf{Efficiency}} & \multirow{2}{*}{\textbf{mAP}} \\
         & & & \textbf{Time} & \textbf{Power} & \textbf{Power} & &  \\ \hline
        \textbf{TensorFlow} & CPU & FP32 & 767.16 ms & N/A & N/A & N/A & 40.7 \\ \hline
        \textbf{TensorRT} & Jetson & FP32 & 69.75 ms & 28 W & 18.5 W & 0.45 $\frac{FPS}{W}$& 39.8 \\
        ~ & Jetson & FP16 & 35.32 ms & 20 W & 10.5 W & 1.17  $\frac{FPS}{W}$ & 39.8 \\
        ~ & Jetson & INT8 & 22.0 ms & 16.5 W & 7.0 W & 2.05  $\frac{FPS}{W}$ & 38.6 \\ \hline
        \textbf{Vitis AI} & ZCU104 & INT8 & 54.42 ms & 23 W & 7.0 W & 1.31  $\frac{FPS}{W}$ & 37.5 \\
        \textbf{Vitis AI Zoo} & ZCU104 & INT8 & 67.84 ms & 22.5 W & 6.5 W & 1.40  $\frac{FPS}{W}$ & 38.3 \\
        \textbf{Zoo Pruned }& ZCU104 & INT8 & 48.42 ms & 21.6 W & 5.6 W & 1.97  $\frac{FPS}{W}$ & 36.5 \\ 
    \end{tabular}
    \caption{Comparison of the optimized models across all used frameworks using all available quantization modes.}
    \label{tab:results}
\end{table}
In addition, a lower energy consumption is evident. The INT8 quantization causes a loss of accuracy compared to the FP32 and FP16 models, but the average inference time was reduced. The Jetson AGX Xavier offers the possibility to configure the maximum clock rate, and therefore the maximum power consumption of the device, through so-called Jetson Power Modes. Mode 1, 2, and 5 budget the maximum power of the Jetson to 10, 15, and 30 watts, respectively. Mode 0 or mode MAXN is selected by default and sets the maximum possible power of the Jetson. The INT8 model optimized with TensorRT is used for the experiment as it shows the highest energy efficiency and by adjusting the power mode, this can be further optimized. As shown in Table~\ref{tab:power-modes-runtime}, the power modes have a direct impact on the overall runtime of the object detection application. Please note that there is a time difference for the same experiment reported in Table~\ref{tab:power-modes-runtime} with 70.23\,ms and in Table~\ref{tab:results} with 69.75\,ms since we run the same experiment multiple times.
\begin{table}
    \centering
    \setlength{\tabcolsep}{10pt}
    \renewcommand{\arraystretch}{1.3}
    \begin{tabular}{c|c|c|c|c}
        ~               & \textbf{MAXN}      & \textbf{30W}      & \textbf{15W}        & \textbf{10W} \\ \hline
        \textbf{Pre-Process}     & 8.78 ms    & 10.91 ms & 16.43 ms   & 17.11 ms \\
        \textbf{Inference}       & 22.0 ms   & 26.22 ms & 33.75 ms   & 61.76 ms \\ 
        \textbf{Post-Process}    & 39.45 ms  & 48.62 ms & 68.12 ms   & 71.71 ms \\ \hline
        \textbf{Total}           & 70.23 ms  & 85.75 ms & 118.3 ms   & 150.58 ms \\
    \end{tabular}
    \caption{Runtime on Nvidia Jetson in different power modes.}
    \label{tab:power-modes-runtime}
\end{table}

For the optimization with Vitis AI, the Vitis AI Quantizer was applied first. The quantized model is also executed on the desktop PC and evaluated with respect to accuracy and inference time. Figure~\ref{fig:map-vai} shows the accuracy loss after all optimization steps for Vitis AI. It was found that the Vitis AI Quantizer reduces the mAP by 2.2\,\%. This result can be verified by checking against reference instructions for optimizing a YOLOv4 model as similar results are obtained. Compared to the baseline model, a significantly higher runtime of the quantized model was observed on the desktop PC. However, it is noticeable that only one processor core is used when executing the model on the Desktop PC. When executing the baseline model, all available cores are used. Thus, the inference of the Vitis AI quantized model is not optimally parallelized. After compiling the model, using the Vitis AI compiler, it is subsequently installed on the ZCU104 and executed by the evaluation framework. It can be seen that the mAP is lower in comparison with the base model, but also inference time is reduced to 54.4\,ms (see Table~\ref{tab:results}. For the optimization via Vitis AI Optimizer, already optimized models from the Vitis AI Model Zoo~\cite{vai_repo} are used and their results are evaluated. We utilize a YOLOv4 model trained on COCO, which is provided pruned and unpruned with an input resolution of $416 \times 416$. In the information provided about each model in Model Zoo, it is described that the number of operations in the model was reduced by 36\,\% using the Vitis AI Optimizer. The two models are also integrated and evaluated in this experiment.
When comparing the similarly optimized models in different input resolutions, it can be seen that the optimizations of the model in different resolutions result in similar improvements in terms of inference time and energy efficiency. Thus, after applying the optimizations, similar results are obtained as shown in Fig.~\ref{fig:map-vs-infer} and Fig.~\ref{fig:map-vs-eff}. No improvements in terms of power consumption could be identified, as shown in Fig.~\ref{fig:map-vs-watt}. It can be concluded that the input resolution has no direct influence on the effectiveness of the applied optimization tools. If a low inference time at an acceptable cost of accuracy is required, a smaller input resolution is preferable.

\subsection{Discussion}
\label{sec:discussion}
\begin{figure}[ht]
\centering
    \begin{subfigure}{0.49\textwidth}
        \includegraphics[width=\textwidth]{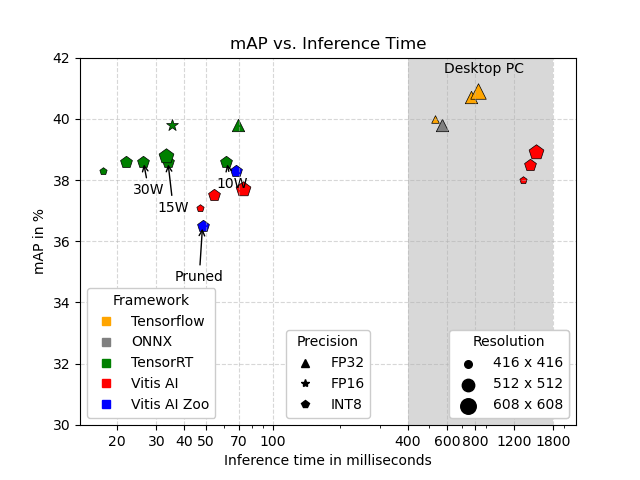}
        \caption{Accuracy vs. Inference Time}
        \label{fig:map-vs-infer}
    \end{subfigure}
    \hfill
    \begin{subfigure}{0.49\textwidth}
        \includegraphics[width=\textwidth]{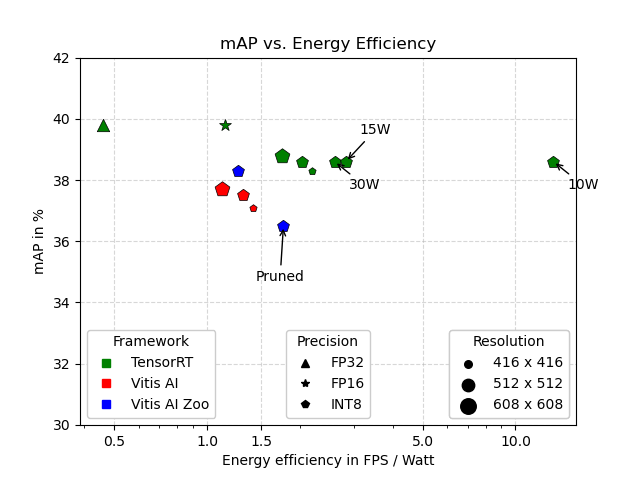}
        \caption{Accuracy vs. Energy Efficiency}
        \label{fig:map-vs-eff}
    \end{subfigure}
    \begin{subfigure}{0.49\textwidth}
        \includegraphics[width=\textwidth]{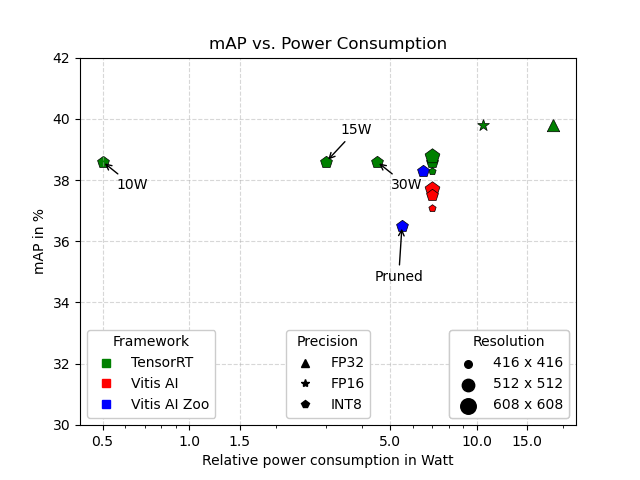}
        \caption{Accuracy vs. Power Consumption}
        \label{fig:map-vs-watt}
    \end{subfigure}
    \caption{Comparison of all optimized models with respect to the chosen evaluation measures accuracy, inference time, power consumption and energy efficiency.}
\end{figure}
The results just described show that significant improvements in runtime and energy efficiency can be achieved by optimizing with TensorRT and Vitis AI. However, a small loss of accuracy can be expected for most optimization operations. Since FP16 quantization with TensorRT does not result in any loss of accuracy, it outperforms all other optimized models when the highest possible accuracy is required. The FP16 model already has a lower runtime than all optimizations in Vitis AI as shown in Fig.~\ref{fig:map-vs-infer}. For the INT8 quantization operations, similar accuracy losses could be observed in TensorRT and Vitis AI. Nevertheless, after the compilation process with the Vitis AI compiler, the accuracy is further reduced. Accordingly, all TensorRT models outperform the Vitis AI models in terms of accuracy.
The TensorRT optimized INT8 model with an input resolution of $416 \times 416$ outperforms all other models in terms of inference time at the expense of an acceptable loss of accuracy. Significant improvements in energy efficiency and inference time can be observed with the Vitis AI Optimizer. On the other hand, the largest accuracy loss of all optimizations is noted here. The measurements of the pruned model in terms of energy efficiency are the only ones that provide similarly good results as the TensorRT-INT8 model. Figure~\ref{fig:map-vs-infer}, ~\ref{fig:map-vs-eff} and~\ref{fig:map-vs-watt} also show the measurements at different power budgets in the Jetson. Power budgeting can increase energy efficiency at the expense of inference time. Hence, a power budget of 15 watts is preferable for the requirement of high energy efficiency instead of 10W Mode, since the inference time is only 11.75\,ms higher in comparison to MAXN Mode.

\section{Conclusion}
\label{sec:conclusion}
We described a generic evaluation framework for computer vision algorithms deployed on an embedded device. As an exemplary task we chose object detection. The framework is \emph{flexible} as different vendor-specific optimization pipelines are supported, \emph{scalable} as multiple edge devices can be connected to the host computer and evaluated in parallel, and \emph{light-weight} as basically only some MQTT messages need to be supported (i.e. interpreted and replied) to integrate new devices. We demonstrate the effectiveness of the framework in multiple experiments using reference hardware and optimization software from Nvidia and Xilinx. Performance deterioration during optimization, the effective runtime reduction as well as energy efficiency can be measured simultaneously.

%
%
%
%
\bibliographystyle{splncs04}
\bibliography{egbib}

\end{document}